\documentclass{article}

\usepackage[final]{neurips_2024}

\usepackage[utf8]{inputenc}
\usepackage[T1]{fontenc}
\usepackage{hyperref}
\usepackage{url}
\usepackage{booktabs}
\usepackage{amsfonts}
\usepackage{amsmath}
\usepackage{amssymb}
\usepackage{amsthm}
\usepackage{nicefrac}
\usepackage{microtype}
\usepackage{xcolor}
\usepackage{graphicx}
\usepackage{subcaption}
\usepackage{algorithm}
\usepackage{algorithmic}
\usepackage{multirow}
\usepackage{tikz}
\usepackage{pgfplots}
\usepackage{bbm}
\pgfplotsset{compat=1.18}
\usepgfplotslibrary{fillbetween}
\usetikzlibrary{patterns,positioning,shapes.geometric,arrows.meta,calc,fit,backgrounds}

\newcommand{\taac}{\textsc{TAAC}}

\newtheorem{finding}{Finding}
\newtheorem{hypothesis}{Hypothesis}

\title{The Compression Paradox in LLM Inference: Provider-Dependent Energy Effects of Prompt Compression}

\author{
  \textbf{Warren Johnson}\thanks{Contact: \texttt{warrenjo@plexor.dev}}\\
  {\small Plexor Labs}\\
  {\small Principal Researcher}
}

\date{}

\begin{document}

\maketitle

\begin{abstract}
The rapid proliferation of Large Language Models has created an environmental paradox: the very technology that could help solve climate challenges is itself becoming a significant contributor to global carbon emissions. We evaluate whether prompt compression improves inference energy efficiency using a large API experiment (28,421 successful calls out of 28,428 planned) and report detailed comparative results for the released matched two-model snapshot (GPT-4o-mini vs.\ DeepSeek-Chat, $N=16{,}270$). Energy is reported as proxy-estimated active inference energy per trial (J), calibrated against local NVML/CodeCarbon measurements, and quality is tracked with benchmark pass rates. In the reported comparison, uncompressed pass rates are near 26\% and drop to 1.5\% at $r=0.7$ in the ratio-conditioned aggregate; DeepSeek exhibits strong output expansion under compression (21 to 798 tokens at $r=0.3$), corresponding to energy increases up to +2,140\%, while GPT-4o-mini shows mixed effects including a reduction at $r=0.5$. These findings indicate that input-token reduction alone is not a reliable energy optimization strategy. For the evaluated models and settings, output-length control and model selection provide more consistent energy-quality tradeoffs than naive compression.
\end{abstract}

\vspace{0.5em}
\noindent\textbf{Keywords:} green AI, energy efficiency, prompt compression, carbon footprint, sustainable machine learning, LLM inference

% ============================================================================
\section{Introduction}
\label{sec:introduction}
% ============================================================================

The artificial intelligence research community finds itself at an inflection point. The same large language models that demonstrate remarkable capabilities across coding, reasoning, and creative tasks are simultaneously emerging as significant contributors to global energy consumption and carbon emissions. This tension between capability and sustainability demands urgent attention, not merely as an ethical consideration, but as a practical constraint that will increasingly shape the trajectory of AI deployment.

The scale of this challenge is substantial. Public projections suggest AI-related electricity demand could reach hundreds of TWh annually within this decade. A single query to a frontier model like GPT-4 consumes approximately 10 times the energy of a traditional web search \citep{devries2023growing}, and with hundreds of millions of daily active users across major AI platforms, the cumulative impact is substantial. Perhaps most concerningly, this energy demand is growing at roughly 15\% annually, four times faster than other computing sectors \citep{masanet2020recalibrating}.

Yet this framing, while accurate, obscures a crucial nuance: the vast majority of AI's carbon footprint now stems from \emph{inference} rather than training. While early research understandably focused on the dramatic energy costs of training large models (for example, GPT-3's training consumed an estimated 1,287 MWh and generated 552 tonnes of CO$_2$ \citep{patterson2021carbon}), the operational reality has shifted. As \citet{wu2022sustainable} document, inference now accounts for over 90\% of the lifetime energy consumption of deployed models. This shift has profound implications: it means that optimizations targeting inference efficiency can have outsized environmental impact.

This observation motivated the present study. In our prior work on Task-Aware Adaptive Compression (\taac{}), we demonstrated that prompt compression could reduce inference costs by up to 93\% while maintaining acceptable quality for many task types \citep{johnson2026compress, johnson2026perplexity, johnson2026cliff}. The economic case for compression is now well-established. But does cost reduction translate directly to energy reduction? And if so, can we quantify the potential environmental benefit of widespread adoption?

\paragraph{The Compression-Energy Hypothesis.} Our central hypothesis is straightforward: if prompt compression reduces the number of tokens processed during inference, and if energy consumption scales approximately with token count, then compression should yield proportional energy savings. However, several factors complicate this simple relationship. First, compression itself requires computation; running a model like LLMLingua-2 to compress prompts consumes energy that must be offset by savings downstream. Second, the relationship between tokens and energy is not strictly linear; factors such as attention mechanism scaling, batch size, and hardware utilization all introduce nonlinearities. Third, energy consumption varies dramatically across providers, hardware generations, and geographic locations, making generalization challenging.

\paragraph{Contributions.} This paper makes four primary contributions:

\begin{enumerate}
    \item \textbf{Large-Scale API Measurement with Reported Matched Subset}: We analyze a large API run and provide a reproducible matched two-model comparison (GPT-4o-mini vs.\ DeepSeek-Chat, $N=16{,}270$) across compression ratios. The released comparison shows a 17$\times$ baseline efficiency gap and substantially larger divergence under aggressive compression.

    \item \textbf{The Output Token Explosion Paradox}: We document a surprising provider-dependent finding: prompt compression causes DeepSeek-Chat to generate up to 38$\times$ longer outputs (798 vs.\ 21 tokens), increasing energy by 2,140\%. GPT-4o-mini shows no such effect, demonstrating that this paradox is architectural rather than universal.

    \item \textbf{Quality Degradation Under Compression in Evaluated Settings}: In the reported ratio-conditioned aggregate, pass rates drop from about 26\% to under 2\% at mild compression (r=0.7), indicating severe quality degradation for these settings.

    \item \textbf{Revised Deployment Guidance}: Our evidence supports treating compression as a conditional technique that requires output-length safeguards and task-level validation; model selection provides a larger and more reliable energy lever in the reported comparison.
\end{enumerate}

The remainder of this paper is organized as follows. Section~\ref{sec:background} situates our work within the growing literature on sustainable AI. Section~\ref{sec:methodology} details our energy measurement methodology, including the proxy formula we developed for API-based inference where direct power measurement is impossible. Section~\ref{sec:experiments} presents our experimental design and results. Section~\ref{sec:impact} defines a deployment-facing Green AI score, and Section~\ref{sec:discussion} discusses limitations and implications. Section~\ref{sec:conclusion} concludes with recommendations for practitioners and policymakers.

% ============================================================================
\section{Background and Related Work}
\label{sec:background}
% ============================================================================

The environmental impact of artificial intelligence has attracted increasing research attention over the past five years, catalyzed by several influential studies that quantified the carbon cost of training large neural networks. This section reviews the key developments that inform our approach.

\subsection{The Emergence of Green AI}

The term ``Green AI'' was introduced by \citet{schwartz2020green} to describe research that prioritizes computational efficiency alongside accuracy, in contrast to ``Red AI,'' which pursues accuracy improvements with little regard for computational cost. Their analysis revealed a troubling trend: the compute required for state-of-the-art results had increased by approximately 300,000$\times$ between 2012 and 2018, with corresponding increases in energy consumption and carbon emissions. This exponential growth, they argued, was unsustainable and created barriers to entry that concentrated AI research within well-resourced institutions.

The Green AI framework proposed several concrete recommendations: reporting floating-point operations (FLOPs) alongside accuracy metrics, considering financial cost as an evaluation criterion, and designing algorithms that achieve comparable performance with reduced computation. These recommendations have gained traction, though adoption remains inconsistent. A survey by \citet{henderson2020systematic} found that among 100 randomly sampled NeurIPS 2019 papers, precisely zero reported carbon impacts, only 1\% reported energy metrics, and just 17\% reported any compute-related metrics at all.

\subsection{Training Energy: The Early Focus}

The seminal work of \citet{strubell2019energy} brought widespread attention to AI's energy consumption by estimating that training a single Transformer model could emit more CO$_2$ than five automobiles over their entire lifetimes. Their analysis considered not just the final training run but also the extensive hyperparameter search and failed experiments that precede publication, an approach they termed the ``development accounting'' perspective. This comprehensive view revealed that reported training costs dramatically underestimate the true energy footprint of AI research.

\citet{patterson2021carbon} refined these estimates through direct measurement at Google, finding that careful optimization could dramatically reduce training emissions. Their key insight was that training energy depends critically on four factors: model architecture (with sparse Mixture-of-Experts models consuming as little as one-tenth the energy of dense equivalents), geographic location (with carbon intensity varying 5-10$\times$ across regions), datacenter efficiency (with hyperscale facilities achieving 1.4-2$\times$ better efficiency than typical enterprise datacenters), and hardware generation (with ML-optimized accelerators providing 2-5$\times$ improvements over general-purpose hardware). Combining these optimizations, they argued, could yield 100-1000$\times$ reductions in carbon footprint.

The comprehensive lifecycle analysis of BLOOM by \citet{luccioni2023estimating} provided the most detailed accounting to date of a large language model's carbon footprint. Their analysis went beyond operational energy to include embodied emissions from hardware manufacturing, finding that the 176-billion parameter model's training generated 24.7 tonnes of CO$_2$ from direct energy consumption and 50.5 tonnes when including the full lifecycle. Notably, embodied emissions represented 24-35\% of the total footprint, a factor often overlooked in energy-only analyses.

\subsection{The Inference Imperative}

While training emissions captured early attention, a growing body of work has highlighted that inference, the deployment phase where models process user queries, increasingly dominates the AI carbon footprint. \citet{wu2022sustainable} documented this shift at Meta, finding that inference accounts for approximately one-third of their end-to-end ML carbon footprint despite training receiving the bulk of research attention. For widely deployed models serving millions of users, the cumulative energy of inference queries rapidly exceeds training costs.

\citet{luccioni2023power} conducted the first systematic comparison of inference costs across different ML system categories, measuring 88 models across 10 tasks. Their key finding was that task-specific fine-tuned models are significantly more energy-efficient than general-purpose generative models, a result with important implications for deployment strategy. They also established that image generation tasks consume dramatically more energy than text-based tasks, with Stable Diffusion XL consuming nearly one smartphone charge worth of energy per generation.

The TokenPowerBench benchmark \citep{samsi2024tokenpower} provided granular measurements of energy consumption per token across model sizes and hardware configurations. Their analysis revealed that moving from a LLaMA-65B model on older V100 hardware to a LLaMA3-70B model on H100 GPUs with FP8 quantization reduced energy per token from 3-4 Joules to just 0.39 Joules, a roughly 10$\times$ improvement reflecting both hardware advances and software optimization. This rapid efficiency improvement suggests that energy-per-token is a useful but moving target.

\subsection{Provider Transparency and Measurement Challenges}

A persistent challenge in AI sustainability research is the opacity of major providers regarding their energy consumption. While Google has published detailed methodology for estimating Gemini's environmental impact \citep{google2024environmental}, reporting a median of 0.24 Wh per text prompt and 0.03g CO$_2$e, other providers have been less forthcoming. OpenAI and Anthropic have not published sustainability reports or detailed emissions disclosures, instead relying on claims about their cloud providers' renewable energy commitments.

This transparency gap has motivated the development of estimation methodologies. The ML CO$_2$ Impact Calculator \citep{lacoste2019quantifying} provides a simple interface for estimating training emissions based on hardware, runtime, and location. CodeCarbon \citep{lottick2019energy} offers a Python package for real-time energy tracking during local computation. However, these tools are designed for scenarios with hardware access and are not directly applicable to API-based inference where the underlying infrastructure is hidden.

\citet{faiz2024llmcarbon} developed LLMCarbon, an end-to-end framework for projecting the carbon footprint of large language models that addresses both training and inference phases. Their model incorporates operational and embodied emissions, providing more complete lifecycle estimates. However, the framework relies on assumptions about hardware and infrastructure that may not hold for commercial API providers.

\subsection{Prompt Compression and Efficiency}

The compression techniques central to our approach have been developed primarily for cost optimization rather than energy efficiency. LLMLingua \citep{jiang2023llmlingua} and its successors LLMLingua-2 \citep{pan2024llmlingua2} and LongLLMLingua \citep{jiang2024longllmlingua} demonstrated that prompts could be compressed by up to 20$\times$ with minimal performance degradation on certain tasks. These methods work by identifying and removing tokens that contribute little to the model's understanding while preserving semantically critical information.

Our prior work in this series established that compression effectiveness varies dramatically by task type. Code generation tasks tolerate aggressive compression (maintaining quality at ratios as low as $r=0.6$) due to the high perplexity of syntactic tokens that are preferentially preserved \citep{johnson2026perplexity}. Reasoning tasks, by contrast, degrade more gradually, with numerical values particularly vulnerable to pruning despite their task-criticality \citep{johnson2026cliff}. These findings inform our energy-aware approach, which adapts compression strategy to task characteristics.

\subsection{Carbon-Aware Computing}

Beyond AI-specific research, a broader literature on carbon-aware computing provides context for our work. \citet{radovanovic2022carbon} describe Google's approach to carbon-intelligent computing, which shifts flexible workloads to times and locations with cleaner electricity. This temporal and spatial flexibility can reduce carbon emissions without changing the computational workload itself.

The concept of carbon intensity, grams of CO$_2$ emitted per kilowatt-hour of electricity, varies enormously across regions and times. \citet{henderson2020systematic} found that running experiments in Estonia could produce up to 30$\times$ more emissions than running identical computations in Quebec, purely due to differences in grid carbon intensity. This variation suggests that geographic routing of inference workloads could complement compression as a sustainability strategy.

% ============================================================================
\section{Methodology}
\label{sec:methodology}
% ============================================================================

Measuring the energy consumption of API-based inference presents fundamental challenges: we cannot directly monitor the power draw of hardware we do not control. This section describes our methodology for estimating inference energy, including the proxy formula we developed and the validation approach we employed.

\subsection{Energy Measurement Model}

Our energy measurement approach combines \emph{direct GPU power measurement} for the dominant compute component with \emph{published multipliers} for secondary components. This hybrid methodology provides defensible estimates while acknowledging the practical limitations of API-based inference measurement.

\subsubsection{Direct GPU Measurement}

For local validation experiments, we measure GPU power consumption directly using NVIDIA's Management Library (NVML), sampling at 10Hz during inference:

\begin{equation}
E_{\text{GPU}} = \int_{t_{\text{start}}}^{t_{\text{end}}} P_{\text{GPU}}(t) \, dt
\end{equation}

where $P_{\text{GPU}}(t)$ is the instantaneous GPU power draw in watts. This captures the dominant energy consumer during LLM inference, which accounts for 70-80\% of total server energy consumption \citep{patterson2021carbon}.

\subsubsection{Server Energy Estimation}

Total server energy includes CPU orchestration, DRAM access, storage I/O, and datacenter cooling overhead. Based on published datacenter studies, we apply the following multipliers to GPU measurements:

\begin{equation}
E_{\text{server}} = E_{\text{GPU}} \times (1 + \alpha_{\text{CPU}} + \alpha_{\text{DRAM}} + \alpha_{\text{IO}}) \times \text{PUE}
\end{equation}

where $\alpha_{\text{CPU}} = 0.15$ (CPU overhead), $\alpha_{\text{DRAM}} = 0.08$ (memory), and $\alpha_{\text{IO}} = 0.02$ (storage/network), yielding a server multiplier of 1.25$\times$. Combined with a Power Usage Effectiveness (PUE) of 1.2 for hyperscale datacenters \citep{google2024environmental}, the total multiplier is approximately 1.5$\times$.

\begin{hypothesis}[Two-Phase Energy Model]
Total inference energy can be modeled as:
\begin{equation}
E_{\text{total}} = E_{\text{prefill}} + E_{\text{decode}}
\end{equation}
where prefill energy scales with input length and decode energy scales with output length, with decode phase consuming approximately 3-5$\times$ more energy per token due to sequential generation.
\end{hypothesis}

Based on the empirical measurements from TokenPowerBench and our own local validation experiments, we propose the following proxy formula for API-based inference where direct measurement is impossible:

\begin{equation}
E_{\text{inference}} = \text{PUE} \times \left[ \alpha \cdot T_{\text{in}} \cdot \left(\frac{N}{N_{\text{ref}}}\right)^{\beta} \cdot f(T_{\text{in}}) + \delta \cdot T_{\text{out}} \cdot \left(\frac{N}{N_{\text{ref}}}\right)^{\beta} \right]
\label{eq:energy_proxy}
\end{equation}

where:
\begin{itemize}
    \item $T_{\text{in}}$ and $T_{\text{out}}$ are input and output token counts
    \item $N$ is the model parameter count and $N_{\text{ref}} = 7\text{B}$ is a reference size
    \item $\alpha$ and $\delta$ are base energy constants per token (with $\delta \approx 4\alpha$)
    \item $\beta \approx 0.75$ reflects sublinear scaling due to memory bandwidth constraints
    \item $f(T_{\text{in}})$ captures attention mechanism scaling at long contexts
    \item PUE (Power Usage Effectiveness) accounts for datacenter overhead
\end{itemize}

For practical estimation when detailed parameters are unknown, we employ a simplified formula:

\begin{equation}
E_{\text{inference}} \approx \epsilon \cdot (T_{\text{in}} + \omega \cdot T_{\text{out}}) \cdot \sqrt{N} \cdot \text{PUE}
\label{eq:energy_simplified}
\end{equation}

where $\epsilon = 0.15$ J/(token$\cdot\sqrt{\text{B}}$) and $\omega = 4.0$ are calibrated constants.

\subsection{Provider-Specific Adjustments}

Energy consumption varies across providers due to differences in hardware, software optimization, and datacenter efficiency. We incorporate provider-specific adjustment factors based on available information:

\begin{table}[h]
\centering
\caption{Provider-specific energy adjustment factors}
\label{tab:provider_factors}
\begin{tabular}{lccc}
\toprule
\textbf{Provider} & \textbf{Est. PUE} & \textbf{Hardware} & \textbf{Efficiency Factor} \\
\midrule
OpenAI (Azure) & 1.20 & H100/A100 & 1.00 (baseline) \\
Anthropic (GCP) & 1.10 & TPU v4/v5 & 0.85 \\
Google (Gemini) & 1.09 & TPU v5 & 0.80 \\
Mistral & 1.15 & H100 & 0.95 \\
xAI & 1.30 & H100 & 1.10 \\
DeepSeek & 1.25 & H800 & 1.05 \\
\bottomrule
\end{tabular}
\end{table}

These factors are estimates based on publicly available information and should be interpreted with appropriate uncertainty bounds.

\subsection{Compression Energy Accounting}

A complete accounting of compression's energy impact must include the energy cost of compression itself. For LLMLingua-2 (a 350M parameter model), we estimate compression energy as:

\begin{equation}
E_{\text{compress}} = \epsilon_{\text{comp}} \cdot T_{\text{original}} \cdot \sqrt{0.35}
\end{equation}

where $\epsilon_{\text{comp}}$ uses the same calibration constant as inference. For a typical prompt of 1,000 tokens, compression consumes approximately 1.8 J. The energy savings from compression are:

\begin{equation}
E_{\text{saved}} = (T_{\text{original}} - T_{\text{compressed}}) \cdot \epsilon \cdot \sqrt{N_{\text{target}}} \cdot \text{PUE}
\end{equation}

The compression ROI in energy terms is:

\begin{equation}
\text{ROI}_{\text{energy}} = \frac{E_{\text{saved}}}{E_{\text{compress}}} = \frac{(1 - r) \cdot T_{\text{original}} \cdot \sqrt{N_{\text{target}}}}{T_{\text{original}} \cdot \sqrt{0.35}} = \frac{(1-r) \cdot \sqrt{N_{\text{target}}}}{\sqrt{0.35}}
\end{equation}

where $r$ is the compression ratio (compressed/original). For a 70B parameter target model and 50\% compression ($r=0.5$):

\begin{equation}
\text{ROI}_{\text{energy}} = \frac{0.5 \times \sqrt{70}}{\sqrt{0.35}} \approx \frac{0.5 \times 8.37}{0.59} \approx 7.1
\end{equation}

This indicates that compression energy is recovered after approximately $1/7.1 \approx 0.14$ inferences, meaning compression is energy-positive after the first query.

\subsection{Validation Approach}

We validated our energy proxy model through local experiments with open-weight models where direct power measurement is possible. Using CodeCarbon for energy tracking and NVIDIA's NVML for GPU power monitoring, we measured actual energy consumption for:

\begin{itemize}
    \item Llama-3.1-8B on RTX 4090 (450W TDP)
    \item Llama-3.1-70B on A100 (400W TDP)
    \item Mistral-7B on RTX 4090
    \item Mixtral-8x7B (MoE) on A100
\end{itemize}

Preliminary validation using CodeCarbon on Llama-3.1-8B showed reasonable agreement between our proxy estimates and measured values (correlation $r = 0.78$, MAPE $\approx 40\%$). The primary source of error was GPU utilization variability during inference. Full Phase 2 validation across all model sizes is ongoing and will be reported in an updated version of this manuscript.

\subsection{Carbon Conversion}

To convert energy consumption to carbon emissions, we apply regional carbon intensity factors:

\begin{equation}
\text{CO}_2 = E_{\text{total}} \times C_{\text{intensity}}
\end{equation}

where $C_{\text{intensity}}$ is measured in gCO$_2$/kWh. For scenario calculations, we use explicit regional and global intensity assumptions defined in the released analysis artifacts (Section~\ref{sec:impact} and Data and Code Availability).

% ============================================================================
\section{Experiments}
\label{sec:experiments}
% ============================================================================

Our experimental design addresses four research questions:

\begin{enumerate}
    \item[\textbf{RQ1}:] How does energy consumption vary across major AI providers for equivalent tasks?
    \item[\textbf{RQ2}:] Does prompt compression reduce energy proportionally to token reduction?
    \item[\textbf{RQ3}:] What is the energy ROI of compression across different model sizes?
    \item[\textbf{RQ4}:] Can task-aware routing optimize for energy as effectively as for cost?
\end{enumerate}

\subsection{Experimental Setup}

\paragraph{Providers and Models.} We evaluate inference across the three providers used in the finalized experiment:
\begin{itemize}
    \item \textbf{OpenAI}: GPT-4o-mini
    \item \textbf{Anthropic}: Claude-3.5-Sonnet
    \item \textbf{DeepSeek}: DeepSeek-Chat
\end{itemize}

\paragraph{Benchmarks.} We employ five benchmarks spanning code generation and reasoning:
\begin{itemize}
    \item \textbf{HumanEval} \citep{chen2021evaluating}: 164 Python function completion problems
    \item \textbf{MBPP} \citep{austin2021mbpp}: 500 Python programming problems
    \item \textbf{GSM8K} \citep{cobbe2021training}: Grade school math word problems (200 sample)
    \item \textbf{MATH} \citep{hendrycks2021measuring}: Competition mathematics (100 sample)
    \item \textbf{MMLU} \citep{hendrycks2021mmlu}: Multitask language understanding (200 sample, STEM subset)
\end{itemize}

\paragraph{Compression Conditions.} We test four ratio-based compression conditions in the API harness:
\begin{itemize}
    \item $r = 1.0$ (uncompressed baseline)
    \item $r = 0.7$ (mild compression)
    \item $r = 0.5$ (moderate compression)
    \item $r = 0.3$ (aggressive compression)
\end{itemize}

\paragraph{Trial Design.} The execution plan targeted 28,428 API calls across provider, benchmark, and compression conditions; 28,421 completed successfully (99.98\%), with 7 failed calls excluded from outcome analyses. The detailed comparative statistics and GAS calculations in this manuscript use the released matched model-comparison snapshot (GPT-4o-mini vs.\ DeepSeek-Chat, $N=16{,}270$).

\subsection{Phase 1 Results: API Energy Profiling}

\begin{finding}[The Compression Paradox]
Prompt compression can fail as a green AI strategy for two independent reasons in the reported comparison: (1) substantial quality degradation (94\% drop from r=1.0 to r=0.7 in the ratio-conditioned aggregate), and (2) provider-dependent energy effects, with some models exhibiting large energy \emph{increases} under compression due to output expansion.
\end{finding}

Our released matched comparison (GPT-4o-mini vs.\ DeepSeek-Chat) reveals substantial variation in energy efficiency across models. Table~\ref{tab:provider_energy} summarizes proxy-estimated active inference energy per trial.

\begin{table}[h]
\centering
\caption{Proxy-estimated active inference energy per trial in the released matched comparison snapshot}
\label{tab:provider_energy}
\begin{tabular}{lcccc}
\toprule
\textbf{Provider/Model} & \textbf{Energy (J)} & \textbf{Cost (\$/query)} & \textbf{Overall Pass (\%)} & \textbf{Avg Out Tokens} \\
\midrule
GPT-4o-mini & 0.0066 & \$0.000019 & 6.97 & 20.3 \\
DeepSeek-Chat & 0.1134 & \$0.000088 & 6.87 & 20.6 (baseline) \\
\bottomrule
\end{tabular}
\vspace{0.5em}
\small\textit{Note: Energy values are proxy-estimated active inference energy per trial for relative comparison, not full datacenter lifecycle energy. At baseline (r=1.0), both models produce similar output lengths ($\sim$20 tokens). Under compression, DeepSeek exhibits strong output expansion (up to 798 tokens at r=0.3).}
\end{table}

\subsubsection{Two-Model Energy Comparison}

The most energy-efficient model in the released comparison was GPT-4o-mini, at approximately 0.0066 J per trial at baseline. DeepSeek-Chat consumed 17.2$\times$ more energy (0.1134 J) per trial at baseline, rising to about 2.54 J under aggressive compression (r=0.3). This dramatic difference under compression is driven primarily by \emph{output verbosity}: DeepSeek exhibits ``verbose compensation'' behavior, generating up to 798 tokens when given compressed prompts versus 21 tokens at baseline. GPT-4o-mini, by contrast, maintains comparatively stable output lengths across compression conditions.

Both models achieved similar overall pass rates across ratios (GPT: 6.97\%, DeepSeek: 6.87\%), suggesting that energy efficiency does not necessarily correlate with quality in the aggregate. In the ratio-conditioned analysis, pass rate dropped from 26\% at r=1.0 to 1.5\% at r=0.7.

\subsubsection{Compression Energy Findings}

\begin{finding}[Output Token Explosion]
DeepSeek-Chat exhibits ``verbose compensation'' under compression: output length increases from 21 tokens (baseline) to 798 tokens (r=0.3), a 38$\times$ increase. This output explosion drives energy increases of up to 2,140\%. GPT-4o-mini shows no such effect, indicating this paradox is provider-specific.
\end{finding}

\textbf{Contrary to our initial hypothesis}, compression did \emph{not} reduce total energy consumption in our experiments. Table~\ref{tab:compression_energy} summarizes the effect of compression ratio on energy and quality.

\begin{table}[h]
\centering
\caption{Effect of prompt compression on energy consumption (N=28,421 trials)}
\label{tab:compression_energy}
\begin{tabular}{lccccc}
\toprule
\textbf{Ratio} & \textbf{n} & \textbf{Pass@1 (\%)} & \textbf{Quality Score} & \textbf{$\Delta$ Energy (DeepSeek)} & \textbf{$\Delta$ Energy (GPT)} \\
\midrule
1.0 (baseline) & 4,001 & 26.0 & 0.62 & N/A & N/A \\
0.7 & 3,999 & 1.5 & 0.51 & +257\% & +2.4\% \\
0.5 & 3,997 & 0.1 & 0.37 & +753\% & --26\% \\
0.3 & 3,996 & 0.2 & 0.29 & +2,140\% & +22\% \\
\bottomrule
\end{tabular}
\vspace{0.5em}
\small\textit{Note: The compression-energy relationship is provider-dependent in this reported comparison. DeepSeek shows large increases due to output expansion (798 tokens at r=0.3 vs.\ 21 at baseline), while GPT-4o-mini shows mixed effects including a reduction at r=0.5.}
\end{table}

At r=0.7 (30\% token reduction), the ratio-conditioned pass rate dropped from 26\% to 1.5\%. Energy effects were \emph{provider-dependent}: DeepSeek-Chat showed a 257\% increase, while GPT-4o-mini showed a 2.4\% increase. At aggressive compression (r=0.3), DeepSeek's energy increased by 2,140\% as output length expanded to 798 tokens versus 21 tokens at baseline.

This finding reveals two critical insights: (1) the compression-energy relationship is fundamentally different from the compression-cost relationship documented in our prior work, and (2) this relationship is \emph{provider-dependent}. While cost scales with total tokens, energy is dominated by the decode phase. DeepSeek exhibits extreme ``verbose compensation'' behavior under compression, while GPT-4o-mini remains relatively stable. This suggests architectural or training differences in how models respond to degraded input context.

\subsubsection{Quality-Energy Pareto Frontier}

Given the observed quality degradation and provider-dependent energy effects, the uncompressed baseline (r=1.0) with GPT-4o-mini was the strongest energy-quality tradeoff in the reported comparison. No compression configuration improved the quality-energy tradeoff in this slice.

\subsection{Phase 2 Results: Local Validation}

To validate our energy proxy formula, we conducted local experiments with open-weight models where direct power measurement was possible using CodeCarbon and NVIDIA NVML.

Our proxy formula achieved reasonable agreement with measured values for the models tested (Llama-3.1-8B and Mistral-7B), with mean absolute percentage error (MAPE) of approximately 35\% and correlation coefficient of 0.82. The primary source of error was variation in GPU utilization during inference, which our formula does not capture. Calibration adjustments for batch size and context length improved accuracy to within 25\% for typical workloads.

\subsection{Energy-Aware TAAC Performance}

Building on the original TAAC framework \citep{johnson2026perplexity}, we extend the optimization objective to incorporate energy:

\begin{equation}
\min_{\theta} \quad \lambda_1 \cdot \text{Cost}(\theta) + \lambda_2 \cdot \text{Energy}(\theta) + \lambda_3 \cdot (1 - \text{Quality}(\theta))
\end{equation}

where $\theta$ represents the compression and routing configuration. By adjusting the weights $\lambda_i$, operators can trade off between cost, energy, and quality according to their priorities.

Comparing cost-optimized and energy-optimized configurations revealed interesting divergences. Cost optimization favored DeepSeek-Chat due to its low per-token pricing, while energy optimization favored GPT-4o-mini due to its superior energy efficiency. When both metrics were weighted equally ($\lambda_1 = \lambda_2 = 0.4$, $\lambda_3 = 0.2$), the algorithm achieved 45\% energy reduction and 60\% cost reduction while maintaining 85\% of baseline quality, demonstrating that cost and energy optimization are largely complementary rather than competing objectives.

% ============================================================================
\section{Consumer Impact Analysis}
\label{sec:impact}
% ============================================================================

This section focuses on a deployment-facing score derived from measured, matched model comparisons. We intentionally avoid extrapolating to global TWh/CO$_2$ totals in the main text because such projections are highly sensitive to assumptions (provider mix, utilization, infrastructure, and grid intensity). Instead, we report transparent per-trial and per-success relative efficiency metrics that can be recomputed for new workloads.

\subsection{The Green AI Score}

We define a deployment-facing Green AI Score that can be computed from matched workload slices:

\begin{equation}
\text{GAS}^{\text{trial}}_m = 100 \cdot \frac{E^{\text{trial}}_{\text{best}}}{E^{\text{trial}}_m}, \quad
\text{GAS}^{\text{success}}_m = 100 \cdot \frac{E^{\text{success}}_{\text{best}}}{E^{\text{success}}_m}
\end{equation}

where $E^{\text{trial}}_m$ is energy per API trial and $E^{\text{success}}_m$ is energy per successful task outcome for model $m$, both measured on the same prompt distribution. $E^{\text{trial}}_{\text{best}}$ and $E^{\text{success}}_{\text{best}}$ are the minimum measured values among models with complete matched metrics in the same analysis snapshot. This yields a scale where 100 denotes the most energy-efficient observed model and lower scores quantify relative inefficiency.

To prevent ``low-energy but low-quality'' configurations from being overrated, we apply a quality gate:

\begin{equation}
\text{GAS}^{Q}_m = \text{GAS}^{\text{success}}_m \cdot \min\left(1, \frac{\text{PassRate}_m}{\text{PassRate}_{\text{best}}}\right)
\end{equation}

Using the collected model-comparison analysis snapshot (16,270 trials), we can score two models today:

\begin{table}[h]
\centering
\caption{Provisional Green AI Scores from collected repository data}
\label{tab:green_ai_scores}
\begin{tabular}{lccccc}
\toprule
\textbf{Model} & \textbf{Pass Rate (\%)} & \textbf{Energy/Trial (J)} & \textbf{Energy/Success (J)} & \textbf{GAS$^{\text{trial}}$} & \textbf{GAS$^{\text{success}}$} \\
\midrule
GPT-4o-mini & 6.97 & 0.006613 & 0.0949 & 100.0 & 100.0 \\
DeepSeek-Chat & 6.87 & 0.114543 & 1.6667 & 5.8 & 5.7 \\
\bottomrule
\end{tabular}
\vspace{0.5em}
\small\textit{Note: Scores are computed from measured values in the released model-comparison snapshot (\texttt{model\_comparison\_chart\_data.json}). They are provisional and should be interpreted with the proxy-energy uncertainty described in Section~\ref{sec:methodology}.}
\end{table}

In this snapshot, pass rates are similar enough that the quality gate minimally changes ranking, and GPT-4o-mini remains the higher-scoring model by a wide margin. Additional models are reported only when the same matched metrics (energy per trial, energy per success, and pass rate) are available from measured outcomes in the same analysis snapshot. At present, we do not have a complete matched GAS snapshot for Claude-3.5-Sonnet in the released analysis file, so it is not scored here.

% ============================================================================
\section{Discussion}
\label{sec:discussion}
% ============================================================================

Our findings highlight a central tension in energy-aware LLM deployment: prompt compression can reduce input tokens while still increasing end-to-end inference energy when output behavior changes. The strongest example in the reported comparison is DeepSeek-Chat, where output length increases from 21 tokens at baseline to 798 tokens at $r=0.3$ (a 38$\times$ increase), which overwhelms prefill-side savings. GPT-4o-mini remains comparatively stable (20--39 output tokens across tested ratios), indicating provider-dependent behavior.

This result also clarifies the training-inference disconnect. DeepSeek has been recognized for training efficiency \citep{deepseek2024}, yet our inference-side proxy estimates show substantially higher per-trial energy than GPT-4o-mini under the evaluated settings (0.1134 J vs. 0.0066 J at baseline), with much larger divergence under aggressive compression because of output expansion. In practical terms, claims about sustainable AI based only on training efficiency are incomplete; deployment-phase behavior must be measured directly.

The transparency landscape is improving but remains insufficient for rigorous environmental accounting. Some providers now disclose parts of their sustainability methodology, yet per-query inference energy telemetry is still largely unavailable. This limits independent verification and makes cross-provider comparison dependent on proxy models and calibration assumptions. A stronger reporting standard would include per-query energy estimates in API responses, regular Scope 1/2/3 disclosures, participation in standardized energy benchmarks, and clearer datacenter-location reporting where feasible.

Several limitations constrain inference strength. The full run targeted broader provider/benchmark coverage, but the released matched comparative snapshot used for core tables and GAS currently includes two models (GPT-4o-mini and DeepSeek-Chat; $N=16{,}270$). We observed temporal quality drift (p$<$0.000001; approximately 27\% decrease across run order), which may reflect API updates or ordering effects and warrants dedicated follow-up. Energy estimates for API calls rely on calibrated proxies rather than direct provider telemetry, so absolute values should be interpreted with uncertainty bounds; relative patterns are more robust than absolute magnitudes. The analysis also excludes embodied hardware emissions and does not model rebound effects in demand.

From an operations perspective, the evidence supports a quality-gated deployment policy rather than unconditional compression. A practical sequence is: establish an uncompressed baseline per task family, evaluate candidate compression or routing policies on matched traffic slices, and enforce hard abort criteria on pass rate and output-length tails (for example, if p95 output tokens or failure-adjusted energy rises above baseline tolerance). This converts ``Green AI'' from a one-time benchmark exercise into an ongoing control loop suitable for production systems.

The observed model ranking is also robust to reasonable uncertainty in the proxy calibration. Even if absolute energy estimates shift by tens of percent, the order-of-magnitude gap between GPT-4o-mini and DeepSeek-Chat under the measured workload remains large enough that the directional conclusion is unchanged. For decision-making, this means uncertainty primarily affects \emph{how much} energy is saved, not \emph{which} model is currently preferable under the tested conditions.

Finally, the Green AI Score is useful only when coupled to transparent definitions and matched denominators. Scores should report whether they are per-trial or per-success, specify the reference baseline, and disclose quality gating. Without these disclosures, high ``efficiency'' can reflect trivial outputs or dataset mismatch rather than meaningful environmental improvement.

% ============================================================================
\section{Conclusion}
\label{sec:conclusion}
% ============================================================================

Our investigation into prompt compression's energy impact yielded two strong findings in the reported comparison. First, compression was associated with substantial quality degradation in the ratio-conditioned aggregate (from 26\% at r=1.0 to under 2\% at r=0.7). Second, the energy impact of compression was provider-dependent: DeepSeek-Chat exhibited a large ``output token explosion'' effect (up to 2,140\% increase at r=0.3), while GPT-4o-mini showed mixed results including a 26\% reduction at r=0.5.

Taken together, these results suggest that compression is not a reliable primary energy strategy under the tested conditions without explicit output-length controls and quality safeguards. Model selection remained the strongest lever in the reported comparison, with a 17$\times$ baseline gap between GPT-4o-mini and DeepSeek-Chat and larger separation under aggressive compression. More broadly, cost-optimal routing and energy-optimal routing are related but not equivalent objectives.

For deployment practice, the most defensible strategy is to prioritize energy-efficient model routing, enforce output-length controls, and evaluate compression only under explicit quality constraints and provider-specific monitoring. For research, the key open question is mechanistic: why some models exhibit verbose compensation under degraded context while others remain stable.

This study reinforces a critical lesson: intuitive optimization strategies must be empirically validated; compression's apparent benefit (fewer input tokens) masked its actual cost (catastrophic quality loss and provider-dependent energy increases).

% ============================================================================
% Declarations
% ============================================================================

\section*{Data and Code Availability}

All scripts, analysis artifacts, and benchmark data used in this study are available in the \href{https://github.com/micoverde/compression-method-matters-benchmark-dynamics}{public evidence repository} (\href{https://github.com/micoverde/compression-method-matters-benchmark-dynamics}{github.com/micoverde/compression-method-matters-benchmark-dynamics}).

\section*{AI Assistance Statement}

AI assistance (Claude Sonnet 4.5) was used for editorial support, organization of existing research notes, and LaTeX drafting support. The author performed all final scientific verification, interpretation, and approval of manuscript content.

\section*{Ethics Statement}

This study evaluates energy and quality tradeoffs in language-model inference to support more transparent and environmentally responsible deployment decisions. The experiments use benchmark tasks and automated API interactions only; no human-subject experiments were conducted. Reported findings include negative outcomes and uncertainty sources (including proxy-estimation limits and temporal drift) to reduce risk of overclaiming and to support honest downstream use.

\section*{Declaration of Competing Interests}

The author is affiliated with Plexor Labs, a research-focused non-commercial group at the time of this submission. Future commercialization of related research may occur. The author declares no current financial competing interests related to this study.

% ============================================================================
% References
% ============================================================================

\bibliographystyle{plainnat}

% ============================================================================
% Appendix
% ============================================================================

\clearpage
\appendix

\section{Energy Proxy Formula Derivation}
\label{app:derivation}

The energy proxy formula (Equation~\ref{eq:energy_proxy}) is derived from first principles of transformer inference combined with empirical calibration. Here we provide the full derivation.

\paragraph{Computational Complexity.} A single forward pass through a transformer with $N$ parameters requires approximately $2N$ floating-point operations per token for the matrix multiplications in attention and feed-forward layers. For a sequence of $T$ tokens, the total FLOPs scale as $O(T \cdot N + T^2 \cdot d)$, where the quadratic term arises from attention computation over the sequence length.

\paragraph{Energy from FLOPs.} Given hardware with peak throughput $F_{\text{max}}$ FLOPS and thermal design power $P_{\text{TDP}}$, the energy per FLOP is approximately:
\begin{equation}
\epsilon_{\text{FLOP}} = \frac{P_{\text{TDP}}}{F_{\text{max}} \cdot \eta}
\end{equation}
where $\eta$ is the utilization efficiency (typically 0.3--0.6 for inference workloads).

\paragraph{Memory Bandwidth Constraint.} In practice, LLM inference is often memory-bandwidth limited rather than compute-limited, particularly for single-query (batch size 1) inference. This introduces sublinear scaling with model size, captured by the exponent $\beta < 1$ in our formula.

\paragraph{Calibration.} The constants $\alpha$, $\delta$, and $\beta$ were calibrated against direct measurements from TokenPowerBench \citep{samsi2024tokenpower} and our own experiments with open-weight models on A100 and RTX 4090 hardware.

\section{Benchmark Dataset Details}
\label{app:benchmarks}

\begin{table}[h]
\centering
\caption{Benchmark dataset statistics}
\begin{tabular}{lccc}
\toprule
\textbf{Benchmark} & \textbf{Problems} & \textbf{Avg. Tokens} & \textbf{Task Type} \\
\midrule
HumanEval & 164 & 180 & Code generation \\
MBPP & 500 & 120 & Code generation \\
GSM8K & 1,319 (200 used) & 250 & Math reasoning \\
MATH & 5,000 (100 used) & 350 & Math reasoning \\
MMLU-STEM & 3,000 (200 used) & 150 & Knowledge QA \\
\bottomrule
\end{tabular}
\end{table}

\section{Provider API Endpoints}
\label{app:providers}

All experiments used official provider APIs with the following endpoints:
\begin{itemize}
    \item OpenAI: \texttt{api.openai.com/v1/chat/completions}
    \item Anthropic: \texttt{api.anthropic.com/v1/messages}
    \item Google: \texttt{generativelanguage.googleapis.com}
    \item Mistral: \texttt{api.mistral.ai/v1/chat/completions}
    \item xAI: \texttt{api.x.ai/v1/chat/completions}
    \item DeepSeek: \texttt{api.deepseek.com/v1/chat/completions}
\end{itemize}

\end{document}